\documentclass[runningheads]{llncs}
\usepackage{graphicx}
\usepackage{tikz}
\usepackage{comment}
\usepackage{amsmath,amssymb} % define this before the line numbering.
\usepackage{color}

\usepackage[width=122mm,left=12mm,paperwidth=146mm,height=193mm,top=12mm,paperheight=217mm]{geometry}

\usepackage{subfigure}
\usepackage{epsfig}
\usepackage{enumitem}
\graphicspath{ {figures/} }
\usepackage[linesnumbered,ruled,vlined]{algorithm2e}
\SetAlFnt{\footnotesize}
\SetKwInput{KwInput}{Input}                % Set the Input
\SetKwInput{KwOutput}{Output}              % set the Output

\usepackage{multirow}
\usepackage{float}
\usepackage{xspace}
\makeatletter
\DeclareRobustCommand\onedot{\futurelet\@let@token\@onedot}
\def\@onedot{\ifx\@let@token.\else.\null\fi\xspace}
\def\eg{\emph{e.g}\onedot} 
\def\ie{\emph{i.e}\onedot} 
\def\cf{\emph{c.f}\onedot} 
 
\def\wrt{w.r.t\onedot} 
\def\etal{\emph{et al}\onedot}
\makeatother

\begin{document}
% \linenumbers
\pagestyle{headings}
\mainmatter
\def\ECCVSubNumber{2348}  % Insert your submission number here

\title{Stochastic Bundle Adjustment for Efficient and Scalable 3D Reconstruction} % Replace with your title

\begin{comment}
\titlerunning{Stochastic Bundle Adjustment for Efficient and Scalable 3D Reconstruction} 
\authorrunning{Lei Zhou, Zixin Luo, Mingmin Zhen, Tianwei Shen, \etal} 
\author{Anonymous ECCV submission}
%\vspace{-1em}
\institute{Paper ID \ECCVSubNumber}
%\vspace{-1.5em}
\end{comment}
%******************

%\begin{comment}
\titlerunning{Stochastic Bundle Adjustment for Efficient and Scalable 3D Reconstruction} 
\author{Lei Zhou\inst{1} \and
Zixin Luo\inst{1} \and 
Mingmin Zhen\inst{1}  \and
Tianwei Shen\inst{1} \and \\
Shiwei Li\inst{2}  \and
Zhuofei Huang\inst{1} \and 
Tian Fang\inst{2} \and
Long Quan\inst{1}
}
\authorrunning{L. Zhou, Z. Luo, M. Zhen, T. Shen, \etal} 
\institute{Hong Kong University of Science and Technology \\
\email{\{lzhouai,zluoag,mzhen,tshenaa,zhuangbr,quan\}@cse.ust.hk} \and
Everest Innovation Technology \\
\email{\{sli,fangtian\}@altizure.com}
}
%\end{comment}
%******************
\maketitle

\begin{abstract}
Current bundle adjustment solvers such as the Levenberg-Marquardt (LM) algorithm are limited by the bottleneck in solving the Reduced Camera System (RCS) whose dimension is proportional to the camera number. When the problem is scaled up, this step is neither efficient in computation nor manageable for a single compute node. 
In this work, we propose a stochastic bundle adjustment algorithm which seeks to decompose the RCS approximately inside the LM iterations to improve the efficiency and scalability. 
It first reformulates the quadratic programming problem of an LM iteration based on the clustering of the visibility graph by introducing the equality constraints across clusters.
Then, we propose to relax it into a chance constrained problem and solve it through sampled convex program.
The relaxation is intended to eliminate the interdependence between clusters embodied by the constraints, so that a large RCS can be decomposed into independent linear sub-problems.
Numerical experiments on unordered Internet image sets and sequential SLAM image sets, as well as distributed experiments on large-scale datasets, have demonstrated the high efficiency and scalability of the proposed approach.
Codes are released at \url{\textit{\textcolor{magenta}{https://github.com/zlthinker/STBA}}}.
\keywords{Stochastic bundle adjustment, Clustering, 3D reconstruction}
\end{abstract}

%%%%%%%%% BODY TEXT

\section{Introduction}
\begin{figure}[t]
\begin{center}
\includegraphics[width=0.45\linewidth]{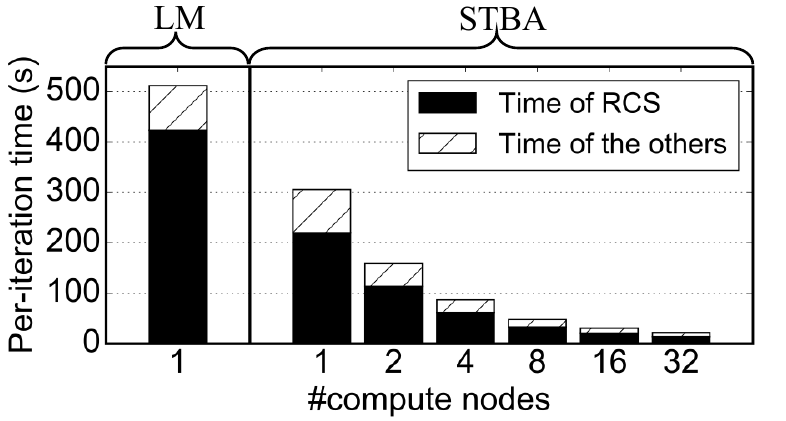}
\end{center}
   \caption{\textbf{Per-iteration time of bundle adjustment \wrt the compute node number.} The Levenberg-Marquardt (LM) algorithm is limited by the bottleneck when solving the reduced camera system (RCS). Our STBA splits the RCS into independent sub-problems, which achieves a speedup on a single-threaded compute node. 
Besides, STBA allows parallel and distributed computing with multiple compute nodes which further improves the efficiency and scalability.}
\label{fig:RCS_timing_bar}
\end{figure}

Bundle Adjustment (BA) is typically formulated as a nonlinear least square problem to refine the parameters of cameras and 3D points.
It is usually addressed by the Levenberg-Marquardt (LM) algorithm, where a linear equation system called \textit{Reduced Camera System} (RCS) \cite{lourakis2009sba,engels2006bundle} must be solved in each iteration.
However, when the problem is scaled up, solving the RCS has been a bottleneck which takes a major portion of computation time (see the first bar of Fig.~\ref{fig:RCS_timing_bar}).
The dimension of the RCS is proportional to the camera number, and thus the increase of cameras would ramp up the computation and memory consumption,
although methods have been proposed to use efficient linear solvers \cite{agarwal2010bundle,kushal2012visibility,dellaert2010subgraph,jian2012generalized,wu2011multicore} and economize on matrix manipulations \cite{agarwal2010bundle,konolige2010sparse,wu2011multicore}.
Furthermore, different to other operations such as Jacobian or gradient evaluations, this step is indivisible, making it hard to fit BA for parallel and distributed computing.

In order to accomplish efficient and scalable reconstructions, clustering has been adopted as a useful practice to decompose a large problem into smaller, more manageable ones.
For example, a number of SfM approaches have been developed in a divide and conquer fashion, which first reconstruct the partitioned sub-maps independently and then merge the partial reconstructions together \cite{ni2007out,zhu2017parallel,zhu2018very,fangmerge}.
Although these methods are able to produce the initial sparse reconstructions in an efficient and scalable way, a full bundle adjustment is still indispensable to optimize the camera and point parameters globally.
Therefore, in the context of BA, the methods \cite{eriksson2016consensus,zhang2017distributed} proposed to distribute the objectives of BA to the split sub-models and optimize the sum of the objectives under the distributed optimization frameworks \cite{combettes2011proximal,bertsekas2014constrained}, which, however, involves extra costly inner iterations and thus makes the optimization over-complicated.

In this work, we follow the direction of exploiting the clustering methods and push forward the investigation on how to integrate a clustering scheme into the BA problem systematically. 
Instead of applying a fixed, and one-time partition at the pre-processing step, we derive a stochastic clustering-based strategy within each LM iteration so as to decompose the RCS for efficiency and scalability.
\begin{itemize}
	\item First, we reformulate the quadratic programming problem of an LM iteration based on the clustering of the visibility graph. Such a formulation splits the problem into the most elementary structures, but meanwhile introduces additional equality constraints and raises the computational cost.
	\item Second, in order to make the above problem efficiently solvable, we propose to relax the constraints into chance constraints \cite{li2008chance} and then solve it with \textit{sampled convex program} \cite{calafiore2005uncertain}. The approach helps to eliminate the interdependence between different clusters by randomized constraint reduction, which hence decomposes the RCS into independent linear sub-problems related to the clusters. In this way, an approximate step can be achieved efficiently.
	\item Third, we present an add-on technique which helps to correct the approximate steps towards the steepest descent direction within a small trust region to improve the convergence.
\end{itemize}
Due to the stochastic process induced by the sampled convex program, we term our algorithm STochastic Bundle Adjustment (STBA), which  brings the following tangible advantages.
First, solving the split RCS in place of the original one has achieved a great speedup thanks to the reduced complexity.
Second, the solving process can be made parallel and scalable to accommodate the growth of camera numbers, 
since all the sub-steps of a BA iteration can be decomposed.
In Fig.~\ref{fig:RCS_timing_bar}, we visualize how the running time is reduced by distributing STBA over multiple compute nodes.

\section{Related Works}

Bundle adjustment (BA) is an indispensable step of 3D reconstruction \cite{zhou2017progressive,luo2018geodesc,luo2019contextdesc,zhou2018learning,zhu2018very,zhu2017parallel}.
It is typically solved by the Levenberg-Marquardt (LM) algorithm \cite{lourakis2005levenberg}, which approximately linearizes the error functions inside a local trust region and then solves a linear \textit{normal equation} for an update step.
SBA \cite{lourakis2009sba} first simplified the norm equation into a \textit{reduced camera system (RCS)} through Schur complement by taking advantage of the special problem structure.
After this, efforts were dedicated to solving the RCS faster in either exact or inexact ways.
The exact solvers apply Cholesky factorization to the reduced camera matrix $\mathbf{S}$, while exploiting variable ordering \cite{amestoy1996approximate,davis2004algorithm} and supernodal methods \cite{rotkin2004design,davis2004algorithm} for acceleration.
The inexact solvers are based on the Conjugate Gradient (CG) method \cite{hestenes1952methods} coupled with various preconditioners \cite{agarwal2010bundle,jeong2011pushing,kushal2012visibility}, which attains inexact solutions with better efficiency.
Apart from the algorithmic improvements, \cite{konolige2010sparse,wu2011multicore} presented well-optimized implementations of the LM solver to save the memory usage and exploit the CPU and GPU parallelism.
However, despite the efforts above, solving a large and indivisible RCS will increasingly become the bottleneck of a BA solver when the problem is scaled up.

In order to make large-scale reconstruction tractable, the clustering methods are initially introduced into the structure from motion (SfM) domain.
Basically, a divide-and-conquer strategy is applied, which first partitions a large scene into multiple sub-maps and then merges the partial reconstructions globally \cite{ni2007out,zhu2017parallel,zhu2018very,fangmerge}.
In the formulation of these approaches, a reduced optimization problem other than the original BA problem is addressed, thus leading to a sub-optimal result.
For example, \cite{ni2007out,zhu2018very} factored out the internal variables inside the sub-maps and \cite{zhu2017parallel} registered all the cameras with motion averaging \cite{chatterjee2013efficient} without the involvement of points.
In the realm of BA, \cite{kushal2012visibility} derived a block-diagonal preconditioner for the RCS from the clustering of cameras, but the clustering did not help to decompose the problem as it is done in the SfM algorithms \cite{ni2007out,zhu2017parallel,zhu2018very,fangmerge}.
Instead, \cite{eriksson2016consensus,zhang2017distributed} proposed to apply the distributed optimization frameworks like the Douglas-Rachford method \cite{combettes2011proximal} and ADMM \cite{bertsekas2014constrained} onto the empirically clusterized BA problems towards large scales.
Although built upon a theoretical foundation, the methods required costly inner iterations and introduced a plethora of latent parameters during optimization.
\iffalse
Rather than solving the RCS with the conventional linear solvers mentioned above, we seek to decompose it into linear sub-problems under the proposed clustering formulation inside the LM iterations.
\fi

\section{Bundle Adjustment Revisited} \label{sec:BA revisited}
In this section, we first revisit the bundle adjustment problem and its LM solution to give the necessary preliminaries and terminologies.
Henceforth, vectors and matrices appear in boldface and $\|.\|$ denotes the L2 norm.

A bundle adjustment problem is built upon a bipartite visibility graph $\mathcal{G} = (\mathcal{C} \cup \mathcal{P}, \mathcal{E})$. 
Here, $\mathcal{C} = \{\mathbf{c}_i \in \mathbb{R}^d\}_{i=1}^m$ denotes the set of $m$ cameras parameterized by $d$-dimensional vectors, $\mathcal{P} = \{\mathbf{p}_i \in \mathbb{R}^3 \}_{i=1}^n$ denotes the set of $n$ 3D points, and $\mathcal{E} = \{\mathbf{q}_i \in \mathbb{R}^2 \}_{i=1}^q$ denotes the set of $q$ projections.
The objective is to minimize $F(\mathbf{x}) =  \| \mathbf{f}( \mathbf{x} ) \|^2$,
where $\mathbf{f}$ denotes a $2q$-dimensional vector of reprojection errors and $\mathbf{x}$ concatenates camera parameters $\mathbf{c} \in \mathbb{R}^{dm}$ and point parameters $\mathbf{p} \in \mathbb{R}^{3n}$, \ie, $\mathbf{x} = [ \mathbf{c}^T, \mathbf{p}^T]^T$. 

The LM algorithm achieves an update step $\Delta\mathbf{x}$ at each iteration by linearizing $\mathbf{f}( \mathbf{x} )$ as $\mathbf{J} (\mathbf{x}) \Delta\mathbf{x} + \mathbf{f}(\mathbf{x})$ in a trust region around $\mathbf{x}$, where $\mathbf{J} (\mathbf{x}) = \mathbf{\nabla}_{\mathbf{x}} \mathbf{f}(\mathbf{x}) = [\mathbf{J} (\mathbf{c}), \mathbf{J} (\mathbf{p})]$ is the Jacobian matrix.
Then the minimization of $F(\mathbf{x})$ is turned into
\begin{equation} \label{eq:quadratic_program}
\min_{\Delta\mathbf{x}} \| \mathbf{J} (\mathbf{x}) \Delta\mathbf{x} + \mathbf{f}(\mathbf{x})  \|^2 + \lambda \|  \mathbf{D} \Delta\mathbf{x} \|^2,
\end{equation}
whose solution comes from the normal equation below
\begin{equation} \label{eq:normal_equation}
\left[
\begin{matrix} 
\mathbf{J} (\mathbf{x}) \\ \sqrt{\lambda} \mathbf{D} 
\end{matrix} 
\right]
\Delta\mathbf{x}
=
\left[
\begin{matrix} 
- \mathbf{f} (\mathbf{x}) \\ \mathbf{0}  
\end{matrix} 
\right],
\end{equation}
where $\lambda > 0$ is the damping parameter and typically $\mathbf{D} = \textrm{diag}(\mathbf{J}^T(\mathbf{x}) \mathbf{J} (\mathbf{x}) )^{\frac{1}{2}}$.
For notational simplicity, we write $\mathbf{J} \triangleq \mathbf{J} (\mathbf{x})$, $\mathbf{J_c} \triangleq \mathbf{J} (\mathbf{c})$, $\mathbf{J_p} \triangleq \mathbf{J} (\mathbf{p})$ and $\mathbf{f}  \triangleq \mathbf{f} (\mathbf{x})$. 
After multiplying $[\mathbf{J}^T, \sqrt{\lambda} \mathbf{D}^T]$ at both sides of Eq.~\ref{eq:normal_equation},
we have
\begin{equation} \label{eq:squared_normal_equation}
(\mathbf{J}^T \mathbf{J} + \lambda \mathbf{D}^T \mathbf{D}) \Delta \mathbf{x} = -\mathbf{J}^T \mathbf{f},
\end{equation}
which can be re-written in the form
\begin{equation} \label{eq:schur}
\left[
\begin{matrix}
\mathbf{B} & \mathbf{E} \\ \mathbf{E}^T & \mathbf{C}
\end{matrix}
\right] 
\left[
\begin{matrix}
\Delta\mathbf{c} \\ \Delta\mathbf{p}
\end{matrix}
\right] 
=
\left[
\begin{matrix}
\mathbf{v} \\ \mathbf{w}
\end{matrix}
\right] ,
\end{equation}
where 
$\mathbf{B} = \mathbf{J}_{\mathbf{c}}^T \mathbf{J_c} + \lambda \textrm{diag}(\mathbf{J}_{\mathbf{c}}^T \mathbf{J_c})$,
$\mathbf{C} = \mathbf{J}_{\mathbf{p}}^T \mathbf{J_p} + \lambda \textrm{diag}(\mathbf{J}_{\mathbf{p}}^T \mathbf{J_p})$,
$\mathbf{E} = \mathbf{J}_{\mathbf{c}}^T \mathbf{J_p}$,
$\mathbf{v} = - \mathbf{J}_{\mathbf{c}}^T \mathbf{f}$, and
$\mathbf{w} = - \mathbf{J}_{\mathbf{p}}^T \mathbf{f}$.
Eq.~\ref{eq:schur} can be simplified by the Schur complement \cite{lourakis2009sba}, which leads to 
\begin{align} 
\mathbf{S} \Delta\mathbf{c} &= \mathbf{v} - \mathbf{E} \mathbf{C}^{-1} \mathbf{w},  \label{eq:camera_update} \\
\Delta\mathbf{p} &= \mathbf{C}^{-1} (\mathbf{w} - \mathbf{E}^T \Delta\mathbf{c}), \label{eq:point_update}
\end{align}
where $\mathbf{S} =  \mathbf{B} - \mathbf{E} \mathbf{C}^{-1} \mathbf{E}^T$ is the Schur complement of $\mathbf{C}$. Here, $\mathbf{S}$, known as the \textit{reduced camera matrix}, is a block structured symmetric positive definite matrix. The block $\mathbf{S}_{ij} \in \mathbb{R}^{d \times d}$ is nonzero iff cameras $\mathbf{c}_i$ and $\mathbf{c}_j$ observe at least one common point.
Although a variety of sparse Cholesky factorization techniques \cite{amestoy1996approximate,davis2004algorithm,rotkin2004design} and preconditioned conjugate gradient methods \cite{agarwal2010bundle,jeong2011pushing,kushal2012visibility} have been developed to solve the \textit{reduced camera system (RCS)} of Eq.~\ref{eq:camera_update}, it still can be prohibitive when the camera number $m$ grows large.

\section{Stochastic Bundle Adjustment} \label{sec:STBA}
In this section, we present our stochastic bundle adjustment (STBA) method that decomposes the RCS into clusters inside the LM iterations.
In Sec.~\ref{sec:clustering_based_reformulation}, we first reformulate problem (\ref{eq:quadratic_program}) based on the clustering of the visibility graph $\mathcal{G}$, yet subject to additional equality constraints.
Next, in Sec.~\ref{sec:chance_constrained_relaxation}, we apply chance constrained relaxation to the reformulation and solve it by sampled convex program \cite{calafiore2005uncertain,campi2011sampling}.
It manages to decompose the RCS into cluster-related linear sub-problems and yield an approximate STBA step efficiently.
Third, a steepest correction step is proposed to remedy the approximation error of the STBA steps in Sec.~\ref{sec:correction}.
Finally,  in Sec.~\ref{sec:clustering}, we present a practical implementation of the random constraint sampler required by the chance constrained relaxation.

\begin{figure}[t]
\begin{center}
\includegraphics[width=1.0\linewidth]{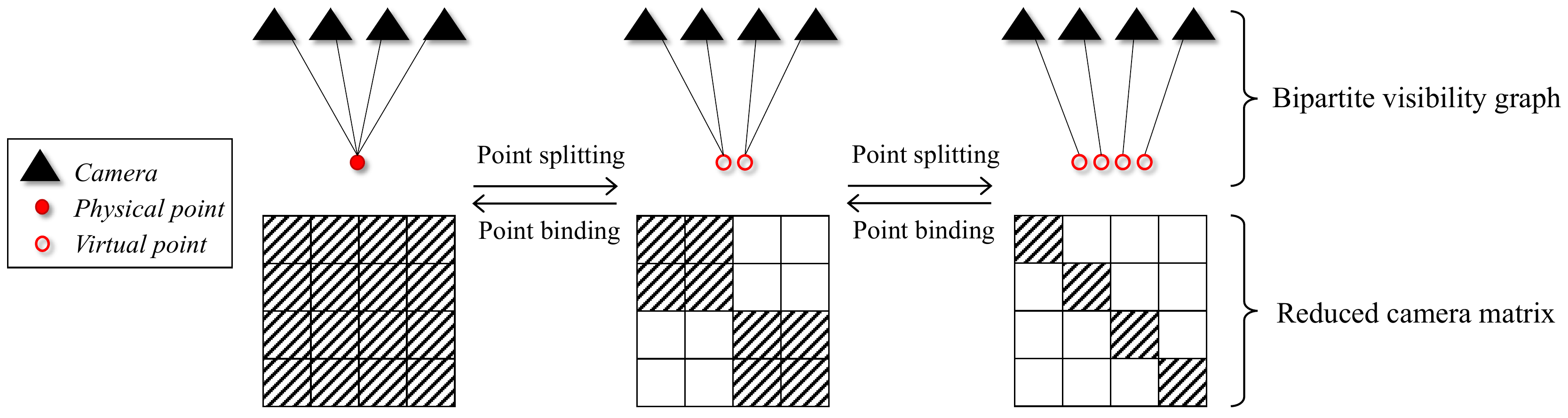}
\end{center}
   \caption{\textbf{Illustration of point splitting/binding over the visibility graph (top) and the corresponding structure of the reduced camera matrix (bottom).} Point splitting helps to reshape the reduced camera matrix into the block-diagonal structure, while point binding does the inverse.}
\label{fig:point_splitting}
\end{figure}

\subsection{Clustering Based Reformulation} \label{sec:clustering_based_reformulation}
In constrast to the previous methods \cite{ni2007out,zhu2017parallel,zhu2018very,fangmerge,eriksson2016consensus,zhang2017distributed} that partition the problem in the pre-processing stage, we present a reformulation of problem (\ref{eq:quadratic_program}) to decompose the RCS into clusters inside the LM iterations.

Particularly, we consider the most general case that every single camera forms a cluster.
In order to preserve all the projections $\mathcal{E}$, we apply \textbf{point splitting} to the physical points, as shown in Fig.~\ref{fig:point_splitting}. For a physical point $\mathbf{p}_i$ viewed by $v_i$ cameras, we split it into $v_i$ virtual points $\{ \mathbf{p}_i^j \}_{j=1}^{v_i}$, each assigned to one cluster.
Such a clustering will reformulate problem (\ref{eq:quadratic_program}) equivalently as a new constrained quadratic programming (QP) problem as below
\begin{align}
\min_{\Delta\mathbf{x}'} & \;\; \| \mathbf{J}' \Delta\mathbf{x}' + \mathbf{f}  \|^2 + \lambda \|  \mathbf{D}' \Delta\mathbf{x}' \|^2, \label{eq:reformulation} \\ 
\textrm{s.t.} & \;\; \mathbf{A} \Delta\mathbf{x}' = \mathbf{0}. \label{eq:constraint}
\end{align}
Here, $\Delta\mathbf{x'} = [\Delta \mathbf{c}^T, \Delta \mathbf{p'}^T]^T$ is an expansion of $\Delta \mathbf{x}$ which considers the update steps for all the virtual points, and so is the Jacobian $\mathbf{J}' = [\mathbf{J}_\mathbf{c}, \mathbf{J}_\mathbf{p}']$.
Accordingly, $\mathbf{D}' = \textrm{diag}(\mathbf{J}'^T \mathbf{J}' )^{\frac{1}{2}}$.
\iffalse
Let us consider the l-th reprojection error $\mathbf{f}_t$ which is associated with a projection from point $\mathbf{p}_i$ to a camera of the k-th cluster $\mathcal{G}_k$. 
By taking the derivative of $\mathbf{f}_t$ \wrt the virtual points of $\mathbf{p}_i$, we have
\begin{equation} \label{eq:jacobian_expansion}
 \frac{\partial \mathbf{f}_l}{\partial \mathbf{p}_i^j} =
\begin{cases} 
\frac{\partial \mathbf{f}_l}{\partial \mathbf{p}_i},  & \text{if } \mathbf{p}_i^j \in \mathcal{G}_k,\\ 
\mathbf{0}, & \text{otherwise}.
\end{cases}
\end{equation}
It indicates that $\mathbf{J}_{\mathbf{p}}'$ is actually obtained by applying zero padding to $\mathbf{J}_{\mathbf{p}}$, and thereby the Jacobian expansion does not introduce extra variables.
\fi
The noteworthy distinction between problems (\ref{eq:reformulation}) and (\ref{eq:quadratic_program}) is that the new equality constraints of Eq.~\ref{eq:constraint} are imposed to enforce that the steps of the same points in different clusters are identical. For example, $\Delta {\mathbf{p}_i^s} = \Delta {\mathbf{p}_i^t} $ $(\forall s, t \in \{1, ..., v_i \})$ for point $\mathbf{p}_i$ and the corresponding j-th row of $\mathbf{A}$ appears in a form as $\mathbf{a}_j = [0..., 1, ..., -1, ...0]$.
Since a point $\mathbf{p}_i$ introduces $v_i - 1$ equations, $\mathbf{A}$ has a row number of $r = \sum_{i=1}^n (v_i-1)$.
Besides, $\mathbf{A}$ is full row rank, because the rows each of which defines a unique equality constraint are linearly independent.

The constrained QP problem above can be easily solved by Lagrangian duality, which turns the problem into 
\begin{equation} \label{eq:clustered_squared_normal_equation}
\mathbf{H}_{\lambda} \Delta\mathbf{x'} = -(\mathbf{J'}^T \mathbf{f} + \mathbf{A}^T \boldsymbol{\nu}),
\end{equation}
where $\mathbf{H}_{\lambda} = \mathbf{J'}^T \mathbf{J'} + \lambda \mathbf{D}'^T \mathbf{D}'$ and $\boldsymbol{\nu} = - (\mathbf{A} \mathbf{H}_{\lambda}^{-1} \mathbf{A}^T)^{-1} \mathbf{A} \mathbf{H}_{\lambda}^{-1} \mathbf{J'}^T \mathbf{f}$ are the Lagrangian multipliers.
Eq.~\ref{eq:clustered_squared_normal_equation} is in the similar format to Eq.~\ref{eq:squared_normal_equation}, but has an additional term $\mathbf{A}^T \boldsymbol{\nu}$ on the right hand side compared with Eq.~\ref{eq:squared_normal_equation}.
While $\mathbf{J'}^T \mathbf{f}$ includes the gradients \wrt the independent virtual points, $\mathbf{A}^T \boldsymbol{\nu}$ acts as a \textbf{correction term} to ensure that the solution complies with the constraints of Eq.~\ref{eq:constraint}.

The ultimate benefit of the clustering-based reformulation is revealed below.
By likewise applying Schur complement to Eq.~\ref{eq:clustered_squared_normal_equation}, we have
\begin{equation} \label{eq:clustered_camera_update} 
\mathbf{S}' \Delta\mathbf{c} = \mathbf{v}' - \mathbf{E}' \mathbf{C}'^{-1} \mathbf{w}',  
\end{equation}
where $\mathbf{E}' = \mathbf{J}_{\mathbf{c}}^T \mathbf{J'_p}$,
$\mathbf{C'} = \mathbf{J'_p}^T \mathbf{J'_p} + \lambda\textrm{diag}(\mathbf{J'_p}^T\mathbf{J'_p})$,
$\mathbf{S}' =  \mathbf{B} - \mathbf{E}' \mathbf{C}'^{-1} \mathbf{E}'^T$,
and $[\mathbf{v}'^T, \mathbf{w}'^T]^T = -(\mathbf{J'}^T \mathbf{f} + \mathbf{A}^T \boldsymbol{\nu})$.
Due to the fact that any two cameras do not share any common virtual points, $\mathbf{S}'$ now becomes a \textbf{block-diagonal} matrix, \ie, $\mathbf{S}'_{ij} = \mathbf{0}, \forall i\neq j$.
Then Eq.~\ref{eq:clustered_camera_update} can be equivalently decomposed into $m$ most elementary linear systems each corresponding to one camera.

\subsection{Chance Constrained Relaxation} \label{sec:chance_constrained_relaxation}
The major problem with the clustering based reformulation above is the excessive cost of evaluating the Lagrangian multipliers $\boldsymbol{\nu}$, because it requires the evaluation of $\mathbf{H}_{\lambda}^{-1}$.
In order to make the problem practically solvable,
we would like to eliminate the need to evaluate the correction term $\mathbf{A}^T \boldsymbol{\nu}$ of Eq.~\ref{eq:clustered_squared_normal_equation} by means of relaxation for problem (\ref{eq:reformulation}).

We multiply a random binary variable $\theta_i$ with the i-th equality constraint of Eq.~\ref{eq:constraint}, which results in $\theta_i \mathbf{a}_i \Delta \mathbf{x}' = 0$.
It could be interpreted that, if $\theta_i = 1$, the constraint $\mathbf{a}_i \Delta \mathbf{x}' = 0$ must be satisfied; otherwise, the constraint is allowed to be violated.
In this way, Eq.~\ref{eq:constraint} will be relaxed into chance constraints \cite{li2008chance}, which leads to
\begin{align}
\min_{\Delta\mathbf{x}'} & \;\; \| \mathbf{J}' \Delta\mathbf{x}' + \mathbf{f}  \|^2 + \lambda \|  \mathbf{D}' \Delta\mathbf{x}' \|^2 , \label{eq:stochastic_reformulation} \\ 
\textrm{s.t.} & \;\; \textrm{Prob}(\mathbf{a}_i \Delta \mathbf{x}' = 0) = \textrm{Prob}(\theta_i = 1) \geq \alpha, \;\;  i = 1, ..., r , \label{eq:stochastic_constraint}
\end{align}
where $\alpha \in (0, 1]$ is a predefined confidence level.
It means that, instead of enforcing the hard constraints, we allow them to be satisfied with a probability above $\alpha$.
The advantage of the chance constrained relaxation is that we can determine the reliability level of approximation by controlling $\alpha$.
The larger $\alpha$ is, the closer the chance constrained problem (\ref{eq:stochastic_reformulation}) will be to the original deterministic problem (\ref{eq:reformulation}).
One approach to problem (\ref{eq:stochastic_reformulation}) is called sampled convex program \cite{calafiore2005uncertain,campi2011sampling}. 
It extracts $N$ independent samples $\Theta(\alpha) = \{\theta_i^{(n)} |i=1,...,r, n=1,...,N  \}$ with a minimum sampling probability of $\alpha$ for each variable $\theta_i^{(n)}$ and replaces the chance constraints (\ref{eq:stochastic_constraint}) with the sampled ones.
%As implied in \cite{calafiore2005uncertain,campi2011sampling}, the larger the sample number $N$ is, the higher the confidence level $\alpha$ will be.
Below we will elaborate on how problem (\ref{eq:stochastic_reformulation}) can be solved given the samples $\Theta(\alpha)$.

For a sample $\theta_i^{(n)} \in \Theta(\alpha)$, if $\theta_i^{(n)} = 0$, the equality constraint $\mathbf{a}_i \Delta \mathbf{x}' = 0$ is dropped;
if $\theta_i^{(n)} = 1$, it enforces the equality of the steps of two virtual points, \eg, $\Delta \mathbf{p}_j^s = \Delta \mathbf{p}_j^t$. Here the virtual point $\mathbf{p}_j^s$ belongs to the single-camera cluster of camera $\mathbf{c}_s$ and similarly $\mathbf{p}_j^t$ to $\mathbf{c}_t$.
Then we merge $\mathbf{p}_j^s$ and $\mathbf{p}_j^t$ into one point as shown in Fig.~\ref{fig:point_splitting}, and we call the operation \textbf{point binding} as opposed to \textbf{point splitting} introduced in Sec.~\ref{sec:clustering_based_reformulation}.
On the one hand, the point binding leads to the consequence that the Lagrangian multiplier $\nu_i = 0$, because the equality $\Delta \mathbf{p}_j^s = \Delta \mathbf{p}_j^t$ always holds.
On the other hand, the block $\mathbf{S}'_{st}$ of $\mathbf{S}'$ (\cf Eq.~\ref{eq:clustered_camera_update}) becomes nonzero, since the merged point is now shared by cameras $\mathbf{c}_s$ and $\mathbf{c}_t$. 
After applying point binding to all the virtual points involved in the sampled constraints, \ie, $\{\theta_i^{(n)} \in \Theta(\alpha) |\theta_i^{(n)}=1 \}$, all the constraints will be eliminated and there is no need to evaluate $\boldsymbol{\nu}$ any more.
Meanwhile, it will bring the cameras sharing common points into the same clusters.

Since the cameras in different clusters have no points in common after the point binding, the matrix $\mathbf{S}'$ will appear in a block-diagonal structure which we call \textbf{cluster-diagonal}, as illustrated in Fig.~\ref{fig:point_splitting}.
It means that each diagonal block of $\mathbf{S}'$ corresponds to a camera cluster. 
In particular, we can intentionally design the sampler $\Theta(\alpha)$ in order to shape $\mathbf{S}'$ into the desired cluster-diagonal structures, as we will present in Sec.~\ref{sec:clustering}.
As a result, this structure of $\mathbf{S}'$ still enables the decomposition of Eq.~\ref{eq:clustered_camera_update} into smaller independent linear systems each relating to one camera cluster.
Provided that there are $l$ clusters, Eq.~\ref{eq:clustered_camera_update} can be equivalently re-written as
\begin{equation} \label{eq:approx}
\begin{cases} 
\mathbf{S}'_1 \Delta \mathbf{c}_1 = \mathbf{b}_1, \\ 
\;\;\;\;\;\; ...  \\ 
\mathbf{S}'_l \Delta \mathbf{c}_l = \mathbf{b}_l,
\end{cases}
\end{equation}
where $[\Delta \mathbf{c}_1^T, ..., \Delta \mathbf{c}_l^T]^T = \Delta\mathbf{c}$ and $[\mathbf{b}_1^T, ..., \mathbf{b}_l^T]^T = \mathbf{v}' - \mathbf{E'} \mathbf{C'}^{-1} \mathbf{w}'$.
After Eqs.~\ref{eq:approx} are evaluated, we substitute $\Delta\mathbf{c}$ into Eq.~\ref{eq:clustered_squared_normal_equation} to give
\begin{equation} \label{eq:cluster_point_update}
\mathbf{J'_p} \Delta \mathbf{p'} = \sum_{i=1}^l \mathbf{J}_{\mathbf{p}^i} \Delta \mathbf{p}^i = -\mathbf{f} - \mathbf{J_c} \Delta \mathbf{c},
\end{equation}
where $\mathbf{J'_p} = [\mathbf{J}_{\mathbf{p}^1}, ..., \mathbf{J}_{\mathbf{p}^l}]$ and
$\Delta \mathbf{p}' = [\Delta \mathbf{p}^1, ..., \Delta \mathbf{p}^l]$ include the Jacobians and virtual point steps \wrt the $l$ clusters respectively and we omit $\mathbf{D}'$ for ease of notation.
To give a uniform step for a physical point, we equalize the steps of its virtual points in different clusters by solving the linear system below in place of Eq.~\ref{eq:cluster_point_update}:
\begin{equation} \label{eq:unified_point_update}
\sum_{i=1}^l \mathbf{J}_{\mathbf{p}^i}   \Delta \mathbf{p} = -\mathbf{f} - \mathbf{J_c} \Delta \mathbf{c}.
\end{equation}
Since $\sum_{i=1}^l \mathbf{J}_{\mathbf{p}^i} = \mathbf{J_p}$, Eq.~\ref{eq:unified_point_update} gives the same solution as the point steps in Eq.~\ref{eq:point_update}: $\Delta\mathbf{p} = \mathbf{C}^{-1} (\mathbf{w} - \mathbf{E}^T \Delta\mathbf{c})$.

So far we have presented how an update step of the camera and point parameters is determined approximately by STBA. Besides this, we keep the other components of the LM algorithm unchanged \cite{marquardt1963algorithm}. For reference, we detail the full algorithm in the supplementary material.

\subsection{Steepest Descent Correction} \label{sec:correction}
The chance constrained relaxation in the last section effectively decomposes the RCS, but leads to approximate solutions with decreased feasibility due to the random constraint sampling.
Below, we provide an empirical analysis on the effect of the approximation and present a conditional correction step to remedy the approximation error.

The LM algorithm is known to be the interpolation of the Gauss-Newton and gradient descent methods, depending on the trust region radius controlled by the damping parameter $\lambda$.
When $\lambda$ is small and the LM algorithm behaves more like the Gauss-Newton method, the approximation induced by STBA in Sec.~\ref{sec:chance_constrained_relaxation} is admissible, in that problem (\ref{eq:quadratic_program}) itself is derived from the first order approximation of the error function $\mathbf{f}(\mathbf{x})$.
And the LM algorithm can automatically contract the trust region when the approximation leads to the increase of the objective.

\begin{figure}[t]
\begin{center}
\includegraphics[width=0.8\linewidth]{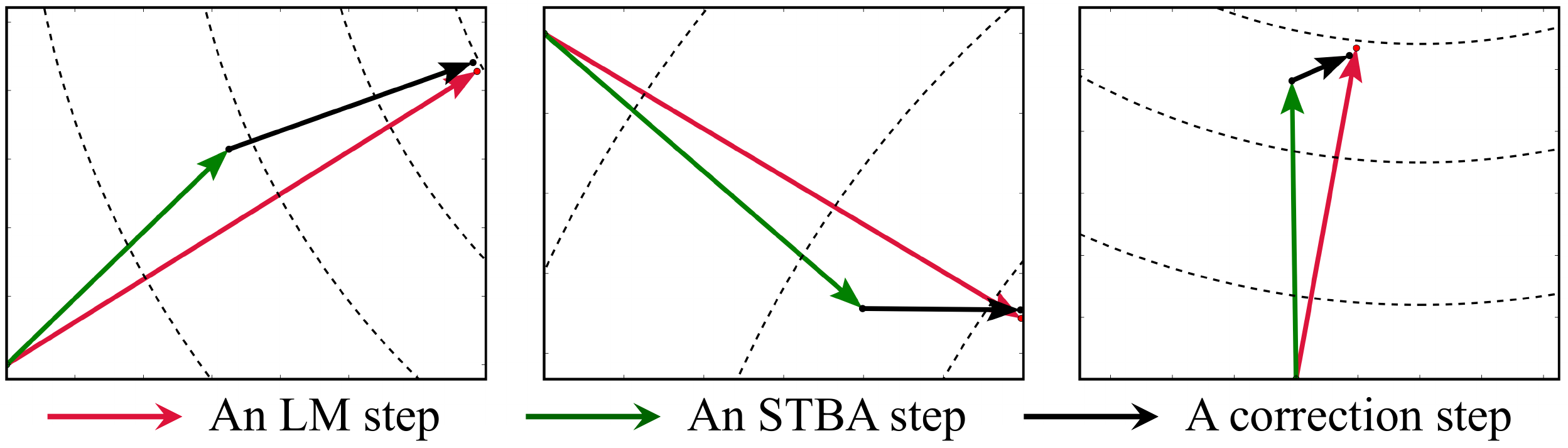}
\end{center}
   \caption{\textbf{Visualization of the steepest descent correction steps} of sample camera parameters when $\lambda = 0.1$. It shows that the deviations of approximate STBA steps from the LM steps are effectively corrected.}
\label{fig:gradient_correction}
\end{figure}

When $\lambda$ is large, \ie, the trust region is small, the LM algorithm is closer to the gradient descent method, which gives a step towards the steepest descent direction defined by the right hand side of Eq.~\ref{eq:clustered_squared_normal_equation}, \ie, $-(\mathbf{J'}^T \mathbf{f} + \mathbf{A}^T \boldsymbol{\nu})$.
However, a problem with STBA is that the correction term $\mathbf{A}^T \boldsymbol{\nu}$ is eliminated approximately by the chance constrained relaxation.
As a consequence, the derived step would deviate from the steepest descent direction and thus hamper the convergence.
Therefore, we propose to recover $\mathbf{A}^T \boldsymbol{\nu}$ to remedy the deviation in such a case.
Especially, when $\lambda$ is large enough, the matrix $\mathbf{H}_{\lambda}$ in Eq.~\ref{eq:clustered_squared_normal_equation} will be dominated by the diagonal terms, so that we can approximate $\mathbf{H}_{\lambda}$ by $\textrm{diag}(\mathbf{H}_{\lambda})$.
After that, $\mathbf{A}^T\boldsymbol{\nu} = - \mathbf{A}^T(\mathbf{A} \mathbf{H}_{\lambda}^{-1} \mathbf{A}^T)^{-1} \mathbf{A} \mathbf{H}_{\lambda}^{-1} \mathbf{J'}^T \mathbf{f}$ can be evaluated efficiently because of the sparsity of $\mathbf{A}$.
Since the approximation $\mathbf{H}_{\lambda} \approx \textrm{diag}(\mathbf{H}_{\lambda})$ is not accurate unless $\lambda$ is large, in practice, we enable the steepest descent correction particularly when $\lambda \geq 0.1$.
Fig.~\ref{fig:gradient_correction} visualizes the effect of the correction.

\subsection{Stochastic Graph Clustering} \label{sec:clustering}

The chance constrained relaxation in Sec.~\ref{sec:chance_constrained_relaxation} necessitates an effective random constraint sampler $\Theta(\alpha)$.
Among many of the possible designs, we propose a viable implementation named stochastic graph clustering in this section.

The design of the clustering method considers the following requirements.
First, the sampler should be randomized with respect to the chance constraints (\ref{eq:stochastic_constraint}). 
Since (\ref{eq:stochastic_constraint}) indicates that the expectation $\mathrm{E}(\theta_i)$ should have $\mathrm{E}(\theta_i) \geq \alpha$ ($i=1,...,r$),
the upper bound of the confidence level $\alpha$ is defined as $\min_{i=1}^r \mathrm{E}(\theta_i)$. 
Therefore, the sampler should sample as many constraints as possible on average to increase the upper bound of $\alpha$. 
Second, the random sampler is intended to partition the cameras into small independent clusters so that Eqs.~\ref{eq:approx} can be solved efficiently.

Concretely, the stochastic graph clustering operates over a camera graph $\mathcal{G}_c = (\mathcal{C}, \mathcal{E}_c)$, where the weight $w_{ij}$ of an edge $e_{ij} \in \mathcal{E}_c$ between cameras $\mathbf{c}_i$ and $\mathbf{c}_j$ is equal to the number of points covisible by the two cameras.
At the beginning, each camera forms an individual cluster as formulated in Sec.~\ref{sec:clustering_based_reformulation}. 
Next, if $\mathbf{c}_i$ and $\mathbf{c}_j$ are joined, a number of $w_{ij}$ pairs of virtual points viewed by $\mathbf{c}_i$ and $\mathbf{c}_j$ will be merged. Therefore, $w_{ij}$ equality constraints will be satisfied.
In order to join as many virtual points as possible while yielding a cluster structure of $\mathcal{G}_c$, we aim at finding a clustering  that maximizes the modularity below inspired by \cite{blondel2008fast}:
$
Q = \frac{1}{2s} \sum_{e_{ij} \in \mathcal{E}_c} \delta(\nu_i, \nu_j) \left(  w_{ij} - \frac{k_i k_j}{2s}  \right),
$
where $s = \sum_{e_{ij} \in \mathcal{E}_c} w_{ij}$ is the total sum of edge weights, $k_i = \sum_{j} w_{ij}$ is the sum of weights of edges incident to camera $\mathbf{c}_i$, and $\nu_i$ denotes the cluster of $\mathbf{c}_i$. $\delta(\nu_i, \nu_j) = 1$ if $\nu_i = \nu_j$ and $0$ otherwise.
The modularity $Q \in [-1, 1]$ measures the density of connections inside clusters as opposed to those across clusters \cite{blondel2008fast}.
Therefore, a larger modularity generally indicates that more virtual points are merged inside clusters.
Maximizing the modularity is NP-hard \cite{schaeffer2007survey}, but Louvain's algorithm \cite{blondel2008fast} provides a greedy strategy which greedily joins the two clusters giving the maximum increase in modularity in a bottom-up manner.
It can be efficiently applied to large graphs since its complexity is shown to be
linear in the node number on sparse data \cite{blondel2008fast}.

\begin{figure}[t]
\begin{center}
\includegraphics[width=0.8\linewidth]{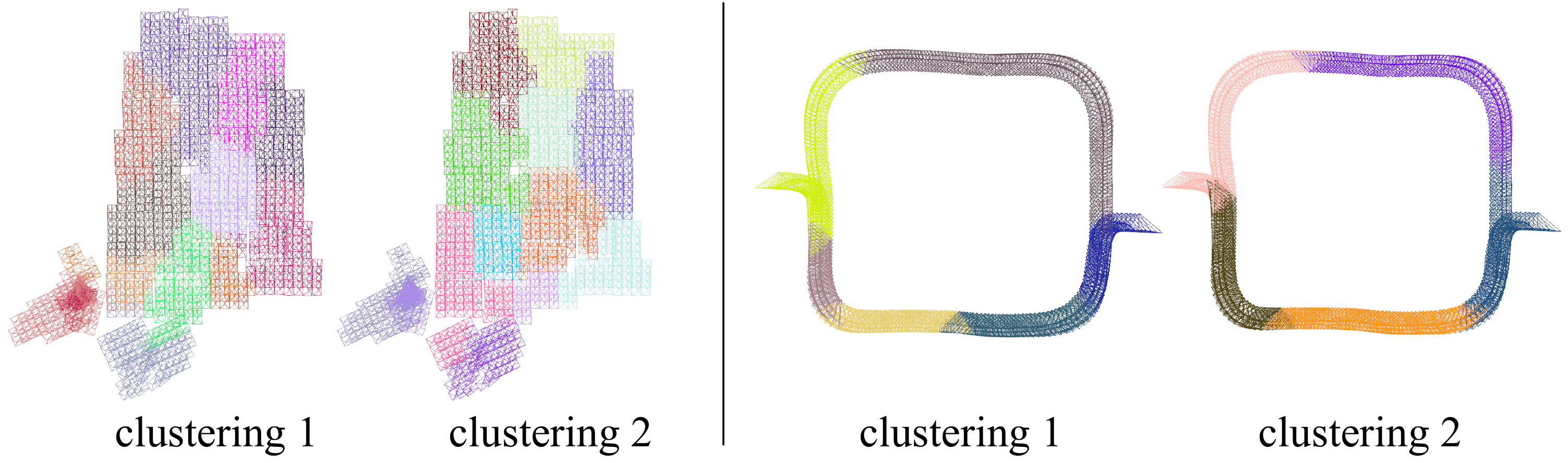}
\end{center}
   \caption{\textbf{Visualization of the random clustering results produced by stochastic graph clustering.} Cameras of different clusters are in different colors.}
\label{fig:louvain}
\end{figure}

However, Louvain's algorithm \cite{blondel2008fast} is deterministic due to its greedy nature. 
To ensure that every pair of virtual points is likely to be merged, we instead join clusters randomly according to a probability distribution defined based on the modularity increments \cite{darmaillac2016mcmc}, which is
\begin{equation} \label{eq:distribution}
\textrm{Prob}(N_x \cup N_y) = \frac{\exp(\beta \Delta Q(N_x, N_y))}{\sum_i \sum_j \exp(\beta \Delta Q(N_i, N_j))},
\end{equation} 
where $\beta>0$ is a scaling parameter.
Two neighboring clusters $N_x$ and $N_y$ are more likely to join together if it leads to a larger modularity increment $\Delta Q(N_x, N_y)$.
In order to limit the sizes of the sub-problems of STBA, we stop joining clusters if their sizes exceed $\Gamma$.
In Fig.~\ref{fig:louvain}, we visualize the stochastic clustering results.

\section{Experiments} \label{sec:experiments}

\subsection{Experiment Settings}

\smallskip\noindent\textbf{Datasets.}
We run experiments on three different types of datasets: 
1)~1DSfM dataset \cite{wilson_eccv2014_1dsfm} which is composed of 14 sets of unordered Internet images;
2)~KITTI dataset \cite{Geiger2012CVPR} containing 11 street-view image sequences;
and 3)~Large-Scale dataset which is collected by ourselves due to the absence of publicly available large-scale 3D datasets. It includes 4 image sets each comprising more than 30,000 images. The problem sizes all exceed the memory of a single compute node and thus we use them particularly for distributed experiments.

\smallskip\noindent\textbf{Comparisons.}
On 1DSfM and KITTI datasets, we compare our method with two standard trust region algorithms, Levenberg-Marquardt (LM) \cite{lourakis2009sba} and Dogleg (DL) \cite{lourakis2005levenberg}.
For the LM algorithm, we use two variants: LM-sparse and LM-iterative, which exploit the exact sparse method and inexact iterative method \cite{kushal2012visibility} to solve the RCS (Eq.~\ref{eq:camera_update}), respectively.
For the distributed experiments on the Large-Scale dataset, we compare our distributed implementation of STBA against the state-of-the-art distributed solver DBACC \cite{zhang2017distributed}.
The ablation studies on steepest descent correction and stochastic graph clustering are presented in Sec.~\ref{sec:ablation_correction} and the supplementary material, respectively.

\smallskip\noindent\textbf{Implementations.}
We implement LM, DL and STBA in C++, using Eigen for linear algebra computations.
All the algorithms are implemented from the same code base, which means that they share the same elementary operations so that they can be compared equitably.
For robustness, we use the Huber loss with a scale factor of 0.5 for the errors \cite{zach2014robust}.
LM-sparse exploits the supernodal $LL^T$ Cholesky factorization with COLAMD ordering \cite{davis2004algorithm} based on CHOLMOD, which is well suited for handling sparse data like KITTI \cite{Geiger2012CVPR}.
LM-iterative uses the conjugate gradient method with the advanced cluster-jacobi preconditioner \cite{kushal2012visibility}.
DL uses the same exact sparse solver as LM-sparse since it requires a reasonably good estimation of the Gauss-Newton step \cite{lourakis2005levenberg}.
Dense $LL^T$ factorization is used to solve the decomposed RCS (Eqs.~\ref{eq:approx}) for STBA due to the dense connectivity inside camera clusters.
Multi-threading is applied to the operations including the reprojection error and Jacobian computation, the preconditioner construction and the matrix-vector multiplications for all the methods as in \cite{wu2011multicore}.
\iffalse
We do not add bells and whistles such as the non-monotonic techniques \cite{deng1993nonmonotonic} onto the implementations, in order to make the comparisons straightforward.
\fi

\smallskip\noindent\textbf{Parameters.}
In the experiments, we assume that the camera intrinsics have been calibrated as in \cite{lourakis2005levenberg,lourakis2009sba}.
Camera extrinsics are parameterized with 6-d vectors, using axis-angle representations for rotations.
We set the initial damping parameter $\lambda$ to 1e-4 and the max iteration number to 100 for all the methods.
The iterations could terminate early if the cost, gradient or parameter tolerance \cite{ceres-solver} drops below 1e-6.
For STBA, we empirically set the scaling parameter $\beta$ to 10 (Eq.~\ref{eq:distribution}) and
the max cluster size $\Gamma$ to 100.

\smallskip\noindent\textbf{Hardware.} We use a compute node with an 8-core Intel i7-4790K CPU and a 32G RAM.
The distributed experiments are deployed on a cluster with 6 compute nodes.

\begin{table}[]
\centering
\caption{\textbf{The statistics of the (a) 1DSfM and (b) KITTI datasets.}}
\subtable[]{
\resizebox{0.47\linewidth}{!}
{
\begin{tabular}{|c|c|c|c|c|}
\hline
Data          & \#images & \#tracks  & \#projections & \#clusters \\ \hline
M. Metropolis & 472      & 73,965    & 528,881       & 6          \\ \hline
M. N. Dame    & 569      & 145,647   & 1,388,635     & 7          \\ \hline
NYC Library   & 577      & 107,867   & 834,298       & 7          \\ \hline
T. of London  & 729      & 166,473   & 1,354,242     & 9          \\ \hline
Ellis Island  & 866      & 167,212   & 1,044,644     & 11         \\ \hline
Alamo         & 890      & 160,652   & 1,937,554     & 11         \\ \hline
P. del Popolo & 1023     & 140,763   & 1,067,134     & 12         \\ \hline
Gen.markt     & 1040     & 213,292   & 1,359,867     & 13         \\ \hline
Yorkminster   & 1057     & 286,372   & 2,083,434     & 13         \\ \hline
V. Cathedral  & 1209     & 303,926   & 2,970,504     & 15         \\ \hline
Roman Forum   & 1799     & 844,159   & 5,815,427     & 22         \\ \hline
Piccadilly    & 3213     & 378,329   & 3,330,024     & 39         \\ \hline
Trafalgar     & 4813     & 337,638   & 3,608,454     & 55         \\ \hline
ArtsQuad      & 5710     & 1,289,493 & 8,676,806     & 65         \\ \hline
\end{tabular}
}
\label{table:1DSfM}
}
\qquad
\subtable[]{
\resizebox{0.4\linewidth}{!}
{
\begin{tabular}{|c|c|c|c|c|}
\hline
Data & \#images & \#tracks & \#projections & \#clusters \\ \hline
00   & 1400     & 119,268  & 475,790       & 20         \\ \hline
01   & 1046     & 96,542   & 577,977       & 13         \\ \hline
02   & 1732     & 161,196  & 591,086       & 25         \\ \hline
03   & 227      & 21,307   & 86,621        & 4          \\ \hline
04   & 153      & 13,081   & 50,762        & 3          \\ \hline
05   & 722      & 63,606   & 246,907       & 10         \\ \hline
06   & 496      & 34,339   & 147,142       & 8          \\ \hline
07   & 254      & 26,160   & 96,607        & 4          \\ \hline
08   & 1205     & 112,435  & 407,985       & 19         \\ \hline
09   & 589      & 55,832   & 204,381       & 9          \\ \hline
10   & 329      & 29,422   & 103,573       & 5          \\ \hline
\end{tabular}
}
\label{table:KITTI}
}
\end{table}

\subsection{Performance Profiles} \label{sec:performance_profile}

Following previous works \cite{dolan2002benchmarking,kushal2012visibility}, we evaluate the solvers with \textit{Performance Profiles} over the total of 25 problems of 1DSfM \cite{wilson_eccv2014_1dsfm} and KITTI \cite{Geiger2012CVPR}.
We obtain the SfM results for 1DSfM by COLMAP \cite{schoenberger2016sfm}\footnote{Since one of the image sets \textit{Union Square} has only 10 reconstructed images, we replace it with another public image set \textit{ArtsQuad}.} and the SLAM results for KITTI by stereo ORB-SLAM2 \cite{murORB2}, while disabling the final bundle adjustment (BA).
Since the SfM/SLAM results are generally accurate because the pipeline uses repeated BA for robust reconstruction, 
we make the problems more challenging by adding Gaussian noise to the points and camera centers following \cite{eriksson2016consensus,jeong2011pushing,ni2007out}.
We report the number of images, tracks and projections of the datasets as well as the typical cluster number of STBA in Table \ref{table:1DSfM} \& \ref{table:KITTI}.

First of all, we give a brief introduction of performance profiles \cite{dolan2002benchmarking}.
Given a problem $p \in \mathcal{P}$ and a solver $s \in \mathcal{S}$, let $F(p, s)$ denote the final objective the solver $s$ attained when solving problem $p$.
Then, for a number of solvers in $\mathcal{S}$, let $F^*(p) = \min_{s \in \mathcal{S}} F(p, s)$ denote the minimum objective the solvers $\mathcal{S}$ attained when solving problem $p$.
Next, we define an objective threshold for problem $p$ which is 
$
F_{\tau}(p) = F^*(p) + \tau (F_0 (p) - F^*(p)),
$
where $F_0 (p)$ is the initial objective and $\tau \in (0, 1)$ is the pre-defined tolerance determining how close the threshold is to the minimum objective.
After this, we measure the efficiency of a solver $s$ by computing the time it takes to reduce the objective to $F_{\tau}(p)$, which is denoted by $T_{\tau}(p, s)$.
And the most efficient solver is the one who takes the minimum time, \ie, $\min_{s \in \mathcal{S}} T_{\tau}(p, s) $.

The method Performance Profiles regards that the solver $s$ solves the problem $p$ if $T_{\tau}(p, s) \leq \alpha \min_{s \in \mathcal{S}} T_{\tau}(p, s)$, where $\alpha \in [1, \infty)$.
Therefore, if $\alpha=1$, only the most efficient solver is thought to solve the problem, while if $\alpha \rightarrow \infty$, all the solvers can be seen to solve the problem. 
Finally, the performance profile of the solver $s$ is defined \wrt $\alpha$ over the whole problem set $\mathcal{P}$ as
$
\rho(s, \alpha) = 100 * \frac{ |\{ p \in \mathcal{P}| T_{\tau}(p, s) \leq \alpha \min_{s\in\mathcal{S}} T_{\tau} (p, s)    \}|}{ |\mathcal{P} |}.
$
It is basically the percentage of problems solved by $s$ and is non-decreasing \wrt $\alpha$.

\begin{figure}[t]
\begin{center}
\includegraphics[width=0.9\linewidth]{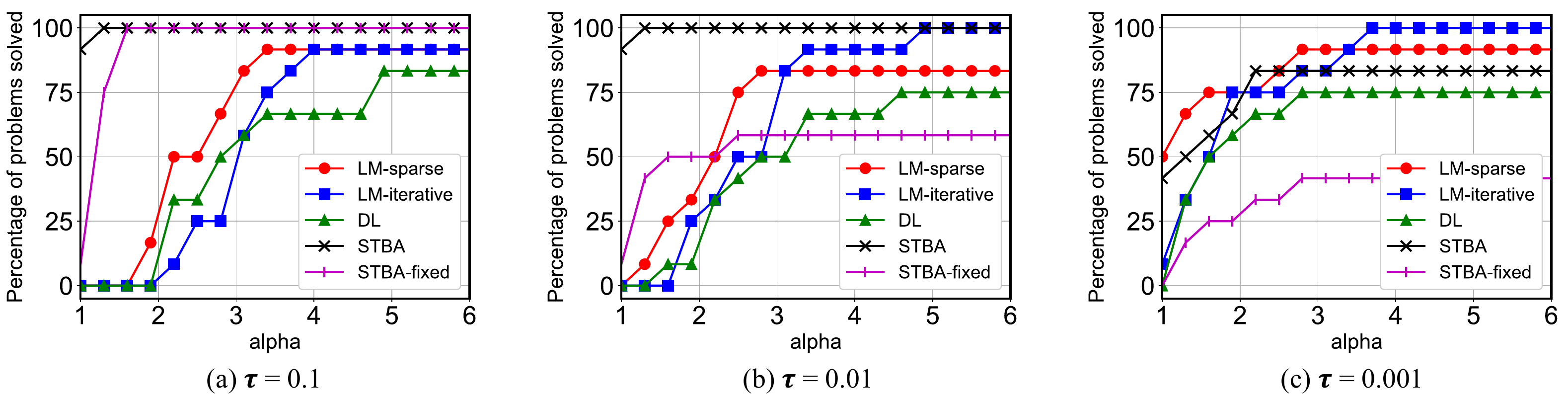}
\end{center}
   \caption{\textbf{Performance profiles \cite{dolan2002benchmarking} of different solvers} when solving the total of 25 problems of 1DSfM \cite{wilson_eccv2014_1dsfm} and KITTI \cite{Geiger2012CVPR}.}
\label{fig:performance_profile}
\end{figure}

\begin{figure*}[t]
\begin{center}
\includegraphics[width=1.0\linewidth]{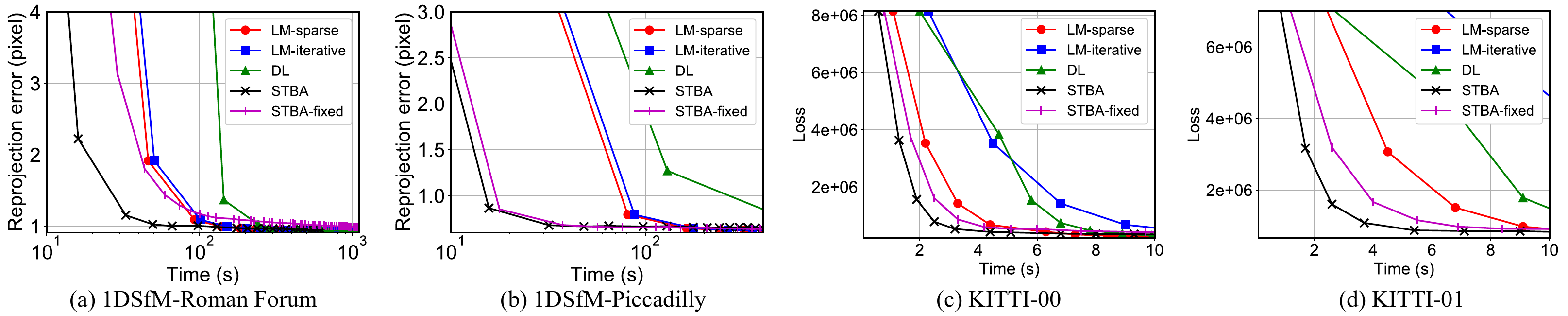}
\end{center}
   \caption{\textbf{The convergence curves of 4 scenes from 1DSfM and KITTI.} }
\label{fig:1DSfM_KITTI}
\end{figure*}

We plot the performance profiles of the solvers in Fig.~\ref{fig:performance_profile}.
To verify the benefits of using stochastic clustering, herein we also compare STBA with its variant STBA-fixed which uses a fixed clustering as previous methods \cite{kushal2012visibility,eriksson2016consensus,zhang2017distributed}.
When $\tau$ is equal to 0.1 and 0.01, our STBA is able to solve nearly 100\% of the problems for any $\alpha$, because it always reaches the objective threshold $F_{\tau}(p)$ with less time than LM and DL methods by a factor of more than 5.
This is mainly attributed to the reduced per-iteration cost as we can see from the convergence curves in Fig.~\ref{fig:1DSfM_KITTI}.
When $\tau$ becomes 0.001 and the threshold $F_{\tau}(p)$ is harder to achieve, STBA is less efficient but still performs on par with LM-iterative and LM-sparse when $\alpha < 3$ and better than DL for any $\alpha$.
On the contrary, the performance of STBA-fixed drops drastically when $\tau$ decreases to 0.001.
The performance change for STBA and STBA-fixed when $\tau$ decreases is mainly caused by the fact that
the clustering methods come with the price of slower convergence near the stationary points \cite{eriksson2016consensus,zhang2017distributed}.
Compared with the full second-order solvers such as LM and DL, clustering methods only utilize the second order information within clusters.
However, as opposed to STBA-fixed which uses fixed clustering, STBA has mitigated the negative effect of clustering by introducing stochasticity so that different second-order information can be utilized to boost convergence as the clustering changes.
Despite the slower convergence rate, the benefit of STBA that it can reduce most of the loss with the lowest time cost (\eg, 99\% loss reduction with only 1/5 time of the counterparts when $\tau=0.01$ in Fig.~\ref{fig:performance_profile}(b)) is still supposed to be highlighted, especially for the real-time SLAM applications where bundle adjustment is called repeatedly to correct the drift.

\subsection{Results on Large-Scale Dataset}

\begin{figure*}[t]
\begin{center}
\includegraphics[width=1\linewidth]{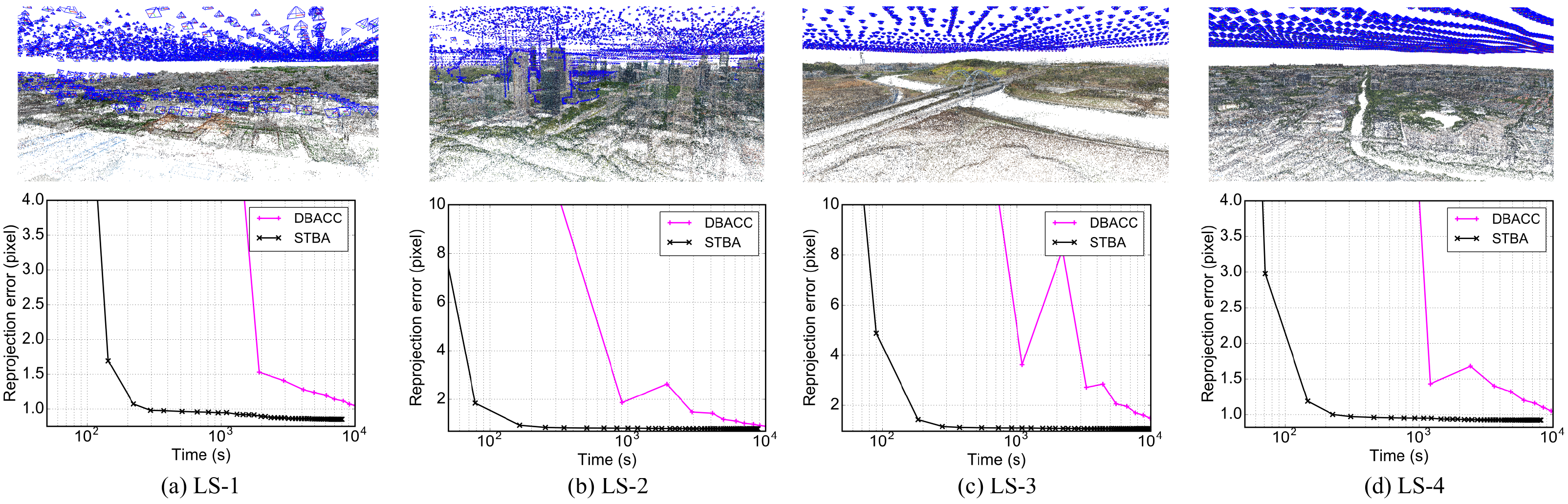}
\end{center}
   \caption{\textbf{Visualizations of SfM results (top) and convergence curves (bottom) of the Larse-Scale dataset.} Cameras are drawn as blue pyramids.}
\label{fig:large_scale}
\end{figure*}

\begin{table}[t]
\centering
\caption{\textbf{Statistics of the distributed bundle adjustment solvers on the Large-Scale dataset.} DBACC \cite{zhang2017distributed} consume many more Jacobian and RCS evaluations than STBA.}
\resizebox{0.8\linewidth}{!}
{
\begin{tabular}{|c|c|c|c|c|c|c|c|c|c|}
\hline
\multirow{2}{*}{Data} & \multirow{2}{*}{\#images} & \multicolumn{2}{c|}{\#clusters} & \multicolumn{2}{c|}{RPE (pixel)} & \multicolumn{2}{c|}{\#Jacobian/RCS evaluations} & \multicolumn{2}{c|}{Mean iteration time (s)} \\ \cline{3-10} 
                      &                           & DBACC           & STBA          & DBACC           & STBA           & DBACC                    & STBA                 & DBACC                  & STBA                \\ \hline \hline
LS-1                  & 29975                     & 300             & 340           & 0.823           & 0.818          & 1011/1080                & 49/100               & 912.5                  & 71.0                \\ \hline
LS-2                  & 33634                     & 336             & 386           & 0.766           & 0.783          & 854/860                  & 48/100               & 934.8                  & 79.3                \\ \hline
LS-3                  & 33809                     & 339             & 391           & 1.083           & 1.056          & 1025/1100                & 49/100               & 1107.0                 & 89.9                \\ \hline
LS-4                  & 44276                     & 444             & 505           & 0.909           & 0.882          & 877/900                  & 49/100               & 988.1                  & 71.2                \\ \hline
\end{tabular}
}
\label{table:large_scale}
\end{table}

To evaluate the scalability, we conduct distributed experiments on the Large-Scale (LS) dataset, which includes the urban scenes of four cities named \textit{LS-1}, \textit{LS-2}, \textit{LS-3} and \textit{LS-4}.
We run a distributed SfM program of our own to produce initial sparse reconstructions of the four scenes and add Gaussian noise with a standard deviation of 3 meters to the camera centers and points.
Then we compare our distributed STBA against the state-of-the-art distributed bundle adjustment framework DBACC \cite{zhang2017distributed}.

We visualize the sparse reconstructions and the convergence curves of the four scenes in Fig.~\ref{fig:large_scale} and report the statistics in Table~\ref{table:large_scale}.
As we can see from Fig.~\ref{fig:large_scale}, STBA achieves faster convergence rates than DBACC by an order of magnitude.
The main cause of the gap is that DBACC, which is based on the ADMM formulation \cite{bertsekas1989parallel}, has to take inner iterations to solve a new minimization problem in every ADMM iteration.
Although we have set the maximum inner iteration number to merely 10, DBACC still takes many more Jacobian and RCS evaluations and thus has much longer iterations than STBA by an order of magnitude, as shown in Table~\ref{table:large_scale}.
Besides, as opposed to DBACC, our STBA is free of too many hyper-parameters.

\subsection{Ablation Study on Steepest Descent Correction} \label{sec:ablation_correction}

\begin{figure*}[t]
\begin{center}
\includegraphics[width=1\linewidth]{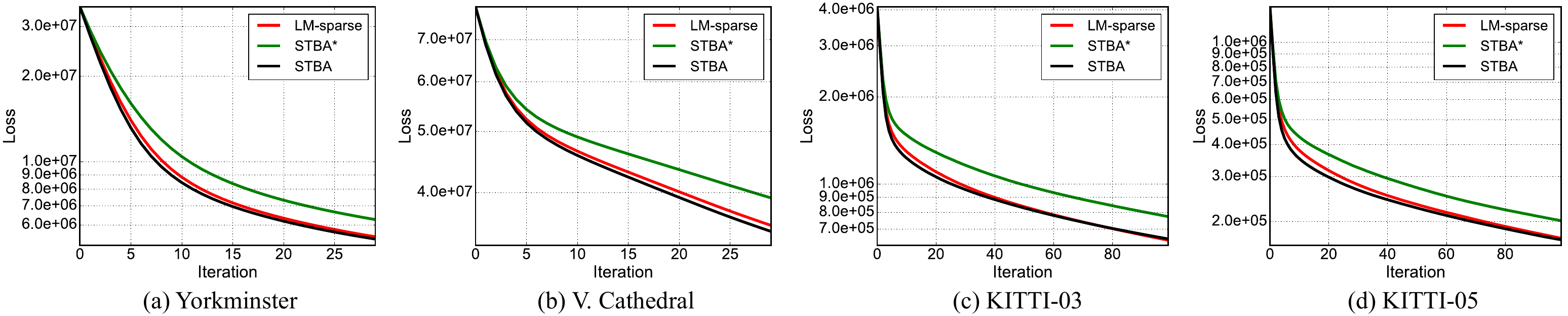}
\end{center}
   \caption{\textbf{Convergence curves of LM-sparse, STBA and STBA* which does not use steepest descent correction (Sec.~\ref{sec:correction}).} Steepest descent correction helps to correct the deviations between the STBA and LM steps.}
\label{fig:gradient_correction_curve}
\end{figure*}

Here we perform an ablation study on steepest descent correction proposed in Sec.~\ref{sec:correction} to validate its efficacy.
We compare our STBA with its variant called STBA* which does not use the correction.
We run STBA, STBA* and LM-sparse on 1DSfM and KITTI.
Since steepest descent correction is designed particularly for a small trust region, we set the lower bound of the damping parameter $\lambda$ to 0.1 in the experiments.
We observe that by using the correction, STBA consistently achieves a faster convergence than STBA* and performs on par with LM-sparse on all the scenes.
Visualizations of the sample convergence curves \wrt the iterations are shown in Fig.~\ref{fig:gradient_correction_curve}, where STBA and LM-sparse have very close convergence curves. 
It manifests that steep descent correction indeed facilitates the correction of the approximation errors of the STBA steps and hence boosts the convergence.

\section{Conclusion} \label{sec:conclusion}

In this paper, we rethink the proper way of integrating the clustering scheme into solving bundle adjustment by proposing STBA.
First, STBA reformulates an LM iteration based on the clustering of the visibility graph, but meanwhile introduces additional equality constraints across the clusters.
Second, we approximately relax the constraints as chance constraints and solve the problem by sampled convex program which randomly samples the chance constraints with the intention of splitting the large reduced camera system into small clusters.
Not only does it reduce the per-iteration cost, but also allows parallel and distributed computing to accommodate the increase of the problem size.
Moreover, we present a steepest descent correction technique to remedy the approximation errors of the STBA steps for a small trust region, and provide a practical implementation of stochastic graph clustering for constraint sampling.
Extensive experiments on Internet SfM data, SLAM data and large-scale data demonstrate the efficiency and scalability of our approach.

\clearpage

\appendix

\noindent{\LARGE \textbf{Appendices}}
\section{Complexity Analysis} \label{sec:complexity}
In this section, we will analyze how our STBA reduces the time and space complexity by solving the split reduced camera system (RCS) (Eqs. \ref{eq:approx}) in place of the original RCS (Eq. \ref{eq:camera_update}), whether the exact or inexact linear solver is used, as shown in Table~\ref{table:complexity}. Please note that the analysis considers the most general case and does not presuppose any special structures, \eg, the extreme sparsity, of the RCS.

\begin{table}[]
\centering
\caption{\textbf{The time and space complexity of LM and our STBA when solving the reduced camera system.} $m$ denotes the camera number. $\Gamma$ is the maximum cluster size. $\kappa$ and $\kappa'$ are the condition numbers of the Schur complement and split Schur complement after preconditioning. $r$ and $r'$ denote the edge number and the sampled edge number of the camera graph, respectively. }
\resizebox{0.5\linewidth}{!}
{
\begin{tabular}{|c|c|c|c|c|}
\hline
                   & \multicolumn{2}{c|}{Cholesky factorization} & \multicolumn{2}{c|}{Conjugate gradient}                  \\ \hline
                   & LM             & STBA              & LM                                       & STBA \\ \hline \hline
Time complexity    & $O(m^3)$   & $O(m\Gamma^2)$ & $O(r\sqrt{\kappa})$ &  $O(r'\sqrt{\kappa '})$             \\ \hline
Space complexity & $O(m^2)$   & $O(m\Gamma)$   & $O(r)$              &    $O(r')$           \\ \hline
\end{tabular}
}
\label{table:complexity}
\vspace{-1.5em}
\end{table}

Cholesky factorization is known to have a cubic time complexity and a quadratic space complexity in the camera number $m$ when solving the RCS \cite{agarwal2010bundle}.
If using Cholesky factorization to solve the split RCS exactly, the time and space complexity of each sub-problem of STBA are $O(\Gamma^3)$ and $O(\Gamma^2)$, respectively, where $\Gamma$ is the maximum cluster size.
Since there are $O(m/\Gamma)$ sub-problems, the time and space complexity of STBA are $O(m\Gamma^2)$ and $O(m\Gamma)$, respectively.
With $\Gamma$ being a constant, the time and space complexity of STBA are linear with $m$.

Besides the exact solvers, conjugate gradient is an inexact approach to solving the linear equations iteratively.
It is known to have a $O(r\sqrt{\kappa})$ time complexity and a $O(r)$ space complexity when solving the RCS \cite{shewchuk1994introduction}, where $r$ is the camera connection number and $\kappa$ is the condition number of the Schur complement $\mathbf{S}$ in Eq. \ref{eq:camera_update}.
However, $\mathbf{S}$ is generally ill-conditioned, which necessitates preconditioning to reduce the condition number $\kappa$ \cite{agarwal2010bundle,kushal2012visibility,dellaert2010subgraph,jian2012generalized}.
The amount of decrease in $\kappa$ depends on how accurately preconditioning can be performed.
If using conjugate gradient to solve the split RCS inexactly, STBA reduces the time complexity to $O(r'\sqrt{\kappa'})$ and the space complexity to $O(r')$. 
Here, $r'$ is the sampled camera connection number, and $\kappa'$ is the maximum condition number of the split Schur complements $\{\mathbf{S}_i \}_{i=1}^l$  after preconditioning.
Due to the sampling of the camera connections, $r'$ is smaller than $r$.
In our experiments, $r'$ is less than one fifth of $r$ when we set the maximum cluster size $\Gamma$ to 100.
The condition number $\kappa'$ also should be smaller than $\kappa$,
because preconditioning the low-dimensional $\mathbf{S}_i$ can be performed more accurately and efficiently than the high-dimensional $\mathbf{S}$.

\section{Ablation Studies on Stochastic Graph Clustering} \label{sec:ablation_clustering}

In Sec.~\ref{sec:clustering}, we have proposed a stochastic graph clustering (SGC) algorithm to sample the chance constraints in each iteration.
In this section, we would like to conduct ablation studies on the clustering strategies and the maximum cluster size $\Gamma$. First, we make comparisons with 3 clustering methods below.

\begin{figure}[t]
\begin{center}
\includegraphics[width=1\linewidth]{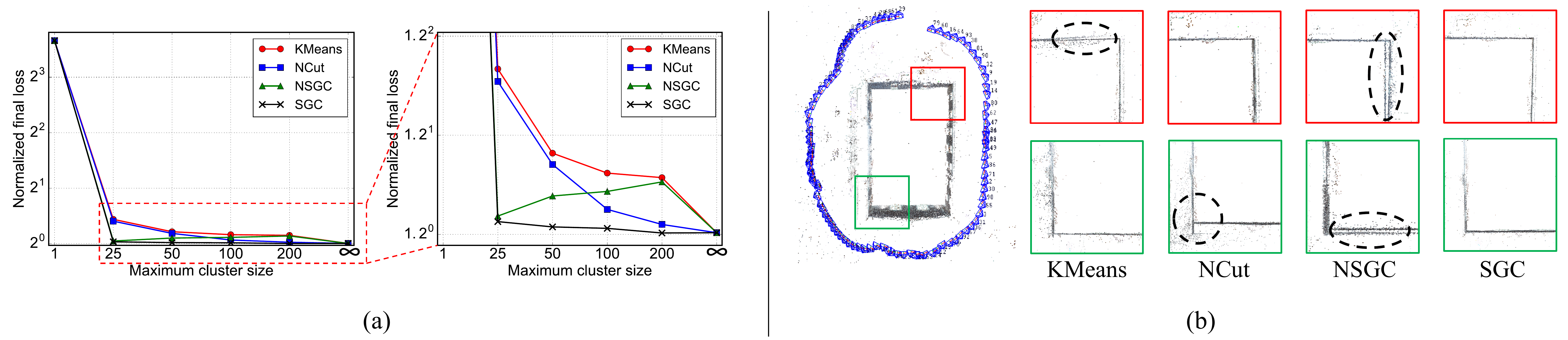}
\end{center}
   \caption{(a) \textbf{Normalized final losses \wrt the maximum cluster size} produced by different clustering methods. The smaller the loss is, the better convergence is attained. (b) \textbf{Reconstructions of Gerrard Hall from the COLMAP dataset \cite{colmap_dataset}.} All the methods except our SGC lead to layered facade reconstruction results, as marked by the dashed circles. (Zoom in for best view.)}
\vspace{-1.5em}
\label{fig:ablation_clustering_example}
\end{figure}

\begin{itemize}[noitemsep,topsep=3pt, leftmargin=6mm] 
\item \textbf{KMeans} which partitions the camera centers into k clusters by using the K-Means algorithm. 
In order to introduce randomness, we randomly choose k camera centers as the initial means in the first step.
\item \textbf{NCut} which uses normalized cut for graph clustering as in the previous works \cite{ni2007out,zhu2017parallel,zhang2017distributed}. 
We turn the camera graph into a random one by keeping its edges with the probability proportional to the edge weights and then run normalized cut on it. 
\item \textbf{NSGC} is the abbreviation for \textbf{non-}stochastic graph clustering. It is a variant of SGC which uses the classic greedy Louvain's algorithm \cite{blondel2008fast} rather than joining clusters randomly as SGC.
\end{itemize}
Apart from the clustering strategies, we run all the algorithms with 6 different maximum cluster sizes which are $\Gamma=$ 1, 25, 50, 100, 200 and $\infty$.
Here, ``$\Gamma=1$" means that each camera forms a cluster.
And ``$\Gamma=\infty$" means all the cameras are grouped into a single cluster, in which case STBA is equivalent to the classic LM algorithm without using clustering.
We run all the methods on each problem of 1DSfM \cite{wilson_eccv2014_1dsfm} and KITTI \cite{Geiger2012CVPR} in the same way as Sec.~\ref{sec:performance_profile}. 
We record the final losses of all the clustering algorithms and normalize them with the division by the minimum loss that the algorithms attained. Therefore, the smaller the normalized loss is, the better convergence is achieved.
We show the average normalized losses of different methods in Fig.~\ref{fig:ablation_clustering_example}(a).

First of all, the proposed SGC reaches the minimum losses at all the cluster sizes, showing its efficacy compared with KMeans and NCut.
The disadvantage of KMeans is that it does not utilize the camera connectivity for clustering, as opposed to NCut and SGC. 
In comparison with SGC, NCut partitions a graph into clusters in a top-down manner.
The downside of this strategy is that it does not explicitly decide whether an edge at the bottom level will be selected or discarded with a defined probability as SGC does (see Eq.~\ref{eq:distribution}).
Since NCut always stops once the cluster sizes are smaller than $\Gamma$, some nodes may constantly stay in the same clusters without being exposed to the cuts.
Instead, the bottom-up strategy of SGC considers the selection of every edge from the very beginning and contributes to the better convergence than NCut in the end.
Besides, the outperformance of SGC over NSGC indicates the necessity of making the graph clustering randomized for better convergence.

Second, all of KMeans, NCut and SGC have better convergence as $\Gamma$ increases.
It is reasonable because the larger $\Gamma$ is, the more chance constraints can be sampled, leading to a more accurate approximation by chance constrained relaxation.
In the extreme case when $\Gamma=1$, all the chance constraints are neglected (\ie, the confidence level $\alpha=0$ in Eq.~\ref{eq:stochastic_constraint}).
It induces poor approximations for the STBA iterations and hence leads to very bad convergence.
However, the final loss can be reduced by an order of magnitude by just increasing $\Gamma$ to 25.
Besides, it is noteworthy that SGC is the least sensitive to $\Gamma$ compared against NCut and KMeans, as the loss does not vary a lot when $\Gamma$ changes from 25 to 200.
Different from other methods, NSGC gets the larger loss when $\Gamma$ increases from 25 to 200.
We found that it is because NSGC uses fixed clusters and neglects the geometric constraints between the clusters all the time, which would cause the inconsistency between the geometries of different clusters.
The problem is more severe when the cluster size increases, as it is less flexible to align large clusters seamlessly than small clusters.
And the reduced flexibility of large clusters is more likely to cause layered geometries at the cluster boundaries (see Fig.~\ref{fig:ablation_clustering_example}(b)).
In Fig.~\ref{fig:ablation_clustering_example}(b), we show the reconstruction results of \textit{Gerrard Hall} from the COLMAP dataset \cite{colmap_dataset} produced by different graph clustering methods with $\Gamma=50$.
All the methods except our SGC lead to layered facade reconstruction results.

\section{Full Algorithm} \label{sec:algorithm}
Below we lay out the full algorithm of STBA.

\begin{algorithm}[]
\footnotesize
\DontPrintSemicolon
  
  \KwInput{Visibility graph: $\mathcal{G} = (\mathcal{C} \cup \mathcal{P}, \mathcal{E})$, initial pose and point parameters: $\mathbf{x} = [\mathbf{c}^T \mathbf{p}^T]^T = \mathbf{x}_0$}
  \KwOutput{$\mathbf{x}^*$ minimizing $F(\mathbf{x})$}
  $t=0$, $t_{max}=100$, $\lambda=1e-4$, $\Gamma=100$, $\epsilon=1e-6$, stop=False\; 
  Build camera graph $\mathcal{G}_c = (\mathcal{C}, \mathcal{E}_c)$ from $\mathcal{G}$\;
  \While{(not stop) and $t++ < t_{max}$}
  {
  \tcc{Stochastic graph clustering}
  $\{\Phi_i\}_{i=1}^l = {\text StochasticGraphClustering}(\mathcal{G}_c, \Gamma)$
  \tcp*{\scriptsize $\Gamma$ is the maximum cluster size}
  Build the equality constraint matrix $\mathbf{A}$ according to $\{\Phi_i\}_{i=1}^l$ \tcp*{\scriptsize see Eq.~\ref{eq:constraint}.}
  \tcc{Evaluations}
  Evaluate reprojection errors $\mathbf{f}$ and Jacobian $\mathbf{J_c}, \mathbf{J_p}, \mathbf{J_p'}, \mathbf{J'} = [\mathbf{J_c}, \mathbf{J_p'}]$\;
  $\mathbf{C} = \mathbf{J_p}^T \mathbf{J_p} + \lambda \textrm{diag}(\mathbf{J_p}^T \mathbf{J_p})$, $\mathbf{E} = \mathbf{J_c}^T \mathbf{J_p}$, $\mathbf{w} = \mathbf{J_p}^T \mathbf{f}$\;
  $\mathbf{B} = \mathbf{J_c}^T \mathbf{J_c} + \lambda \textrm{diag}(\mathbf{J_c}^T \mathbf{J_c})$\;
  $\mathbf{C'} = \mathbf{J_p'}^T \mathbf{J_p'} + \lambda \textrm{diag}(\mathbf{J_p'}^T \mathbf{J_p'})$\;
  $\mathbf{E}' = \mathbf{J_c}^T \mathbf{J_p'}$\;
  $  \mathbf{g} = -\mathbf{J'} \mathbf{f}$  \;
  \tcc{Steepest descent correction}  
\If{$\lambda \geq 0.1$ }
{
$\mathbf{H}_{\lambda} = \mathbf{J'}^T \mathbf{J'} + \lambda \mathbf{D}'^T \mathbf{D}'$ \;
$\mathbf{\tilde{H}}_{\lambda} = \textrm{diag}(\mathbf{H}_{\lambda})$ \;
$\boldsymbol{\nu} = (\mathbf{A} \mathbf{\tilde{H}}_{\lambda}^{-1} \mathbf{A}^T)^{-1} \mathbf{A} \mathbf{\tilde{H}}_{\lambda}^{-1} \mathbf{g}$ \;
$\mathbf{g} = \mathbf{g} - \mathbf{A}^T \boldsymbol{\nu}$
}
 $\mathbf{g} \triangleq [\mathbf{v}'^T \mathbf{w}'^T]^T$ \;
  $\mathbf{S'} = \mathbf{B} - \mathbf{E'} \mathbf{C'}^{-1} \mathbf{E'}^T \triangleq \{\mathbf{S}_i\}_{i=1}^l$ \;
  $\mathbf{b'} = \mathbf{v'} - \mathbf{E'} \mathbf{C'}^{-1} \mathbf{w'} \triangleq \{ \mathbf{b}_i \}_{i=1}^l$\;

  \tcc{Solve pose steps in parallel}
  \For{$i=1$ to $l$ }	
  {
  \textrm{Solve  }$\mathbf{S}_i \Delta\mathbf{c}_i = \mathbf{b}_i$
  }
  $\Delta\mathbf{c} = [\Delta\mathbf{c}_1^T ... \Delta\mathbf{c}_l^T]^T$\;
 
  $\Delta\mathbf{p} = \mathbf{C}^{-1} (\mathbf{w} - \mathbf{E}^T \Delta\mathbf{c})$\;
  $\mathbf{x} = [\mathbf{c}^T \mathbf{p}^T]^T$, $\Delta\mathbf{x} = [\Delta\mathbf{c}^T \Delta\mathbf{p}^T]^T$\;
  \If{(Cost tolerance $<\epsilon$) or (Gradient tolerance $<\epsilon$) or (Parameter tolerance $<\epsilon$) (see \cite{ceres-solver})}
  {
  stop=True
  }
  \If{$F(\mathbf{x}) > F(\mathbf{x} + \Delta\mathbf{x})$}
  {
  $\lambda = \lambda / 3 $, $\mathbf{c} = \mathbf{c} + \Delta\mathbf{c}$, $\mathbf{p} = \mathbf{p} + \Delta\mathbf{p}$\;
  }
  \Else
  {
  $\lambda = \lambda * 3 $\;
  }
  }
$\mathbf{x}^* = [\mathbf{c}^T \mathbf{p}^T]^T$
\caption{\textbf{Stochastic Bundle Adjustment (STBA)}}
\label{algo:STBA}
\end{algorithm}

\clearpage

\bibliographystyle{splncs04}
\bibliography{egbib}
\end{document}